\documentclass[preprint,12pt,authoryear]{elsarticle}
\usepackage{microtype}
\usepackage[T1]{fontenc}
\usepackage{helvet}    
\usepackage{courier}   
\usepackage{booktabs}
\usepackage{tabularx}
\usepackage{subcaption}
\usepackage[normalem]{ulem}
\useunder{\uline}{\ul}{}
\usepackage{enumitem}
\usepackage{amssymb}
\usepackage{amsthm}
\usepackage{amsmath}
\usepackage{lineno}
\usepackage{color}
\usepackage{algorithm}
\usepackage{algpseudocode}
\usepackage{multirow}
\usepackage{array}
\usepackage{hyperref}
\hypersetup{
    colorlinks=true,    
    pdfborder={0 0 0},   
    linkcolor=blue,      
    urlcolor=black,       
    citecolor=blue,      
    filecolor=blue       
}


\usepackage{geometry}
\usepackage{cleveref}
\crefname{table}{Table}{Table}
\crefname{section}{Section}{Section}
\crefname{figure}{Fig.}{Figure}
\crefname{equation}{Eq.}{Eqs.}
\definecolor{green}{RGB}{0, 200, 83}
\journal{}

\begin{document}
\begin{frontmatter}

\title{MoGERNN: An Inductive Traffic Predictor for Unobserved Locations}

\author[a,b]{Qishen Zhou}
\ead{qishenzhou@intl.zju.edu.cn}
\author[c,b]{Yifan Zhang}
\ead{yifan.zhang@cityu-dg.edu.cn}
\author[b]{Michail A. Makridis}
\ead{michail.makridis@ivt.baug.ethz.ch}
\author[b]{Anastasios Kouvelas}
\ead{kouvelas@ethz.ch}
\author[a]{Yibing Wang}
\ead{wangyibing@zju.edu.cn}
\author[d,a]{Simon Hu\corref{cor1}}
\cortext[cor1]{Corresponding Author: Simon Hu}
\ead{simonhu@zju.edu.cn}

\affiliation[a]{organization={Institute of Intelligent Transportation Systems, College of Civil Engineering and Architecture, Zhejiang University},
            city={Hangzhou},
            postcode={310058}, 
            country={China}}

\affiliation[b]{organization={Institute for Transport Planning and Systems, ETH Zurich},
            city={Zurich},
            postcode={8093}, 
            country={Switzerland}}

\affiliation[c]{organization={Department of Computer Science, City University of Hong Kong (Dongguan)},
            city={Dongguan},
            postcode={523808},
            country={China}}
            
\affiliation[d]{organization={ZJU-UIUC Institute, Zhejiang University},
            city={Haining},
            postcode={314400},
            country={China}}

\begin{abstract}
Given a partially observed road network, how can we predict the traffic state of interested unobserved locations? Traffic prediction is crucial for advanced traffic management systems, with deep learning approaches showing exceptional performance. However, most existing approaches assume sensors are deployed at all locations of interest, which is impractical due to financial constraints. Furthermore, these methods are typically fragile to structural changes in sensing networks, which require costly retraining even for minor changes in sensor configuration. To address these challenges, we propose MoGERNN, an inductive spatio-temporal graph model with two key components: (i) a Mixture of Graph Experts (MoGE) with sparse gating mechanisms that dynamically route nodes to specialized graph aggregators, capturing heterogeneous spatial dependencies efficiently; (ii) a graph encoder-decoder architecture that leverages these embeddings to capture both spatial and temporal dependencies for comprehensive traffic state prediction. Experiments on two real-world datasets show MoGERNN consistently outperforms baseline methods for both observed and unobserved locations. MoGERNN can accurately predict congestion evolution even in areas without sensors, offering valuable information for traffic management. Moreover, MoGERNN is adaptable to the changes of sensor network, maintaining competitive performance even compared to its retrained counterpart. Tests performed with different numbers of available sensors confirm its consistent superiority, and ablation studies validate the effectiveness of its key modules. The code of this work is publicly available at: \url{https://github.com/ZJU-TSELab/MoGERNN}.
\end{abstract}

\begin{keyword}
spatio-temporal extrapolation \sep traffic state estimation \sep traffic prediction \sep Kriging \sep inductive graph representation learning \sep mixture of experts
\end{keyword}

\end{frontmatter}

\section{Introduction} \label{Sec:Intro}

\begin{figure}[t]
    \centering
    \includegraphics[width=0.9\linewidth]{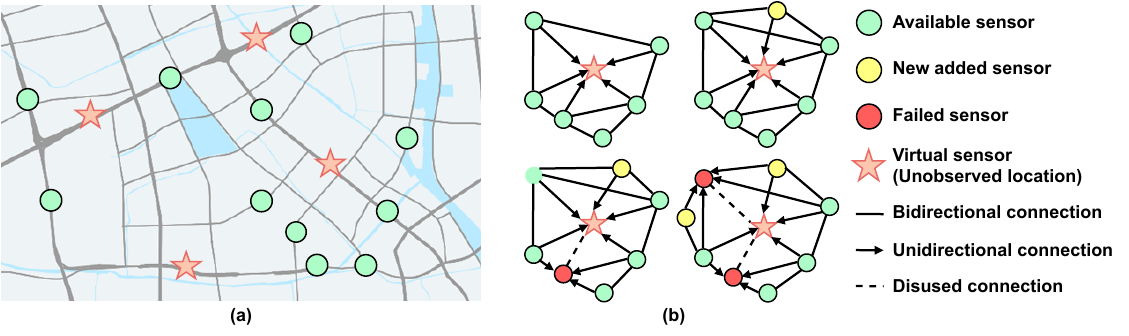}
    \vspace{-0.3cm}
    \caption{An illustration of inductively Forecasting unobserved nodes (iFun), for the sake of presentation, unobserved locations are considered as being equipped with virtual sensors \citep{VirtualSensor2009,VirtualSensor2021}: (a) an example of sensor deployment on the map, with locations of interest without physical sensors; (b) graph typologies for changes in sensor configuration.}
    \label{fig:problem}
\end{figure}

Large-scale traffic forecasting plays a pivotal role in advanced traffic management systems and has gained considerable attention over recent decades. Among the existing research, deep learning-based approaches dominate and achieve extraordinary performance \citep{SHAYGAN2022Review,Yin2022Review}. In particular, Graph Neural Networks (GNNs) have established state-of-the-art results by effectively capturing heterogeneous spatio-temporal dynamics in traffic flow \citep{jin2024Review}. Despite these advances, two critical issues have received limited attention. First, prediction of traffic states at unobserved locations: nearly all existing studies presuppose the presence of sensors at the all locations of interest. However, due to financial constraints, it is impractical to deploy and maintain sufficient sensors across all target areas. Second, fragility to the structural changes of sensing networks: most approaches are transductive, requiring complete model retraining even for minor changes in sensor configuration, such as the addition of a single sensor. In real-world applications, where sensor networks undergo changes due to inevitably sensor retirement and new installations, along with potential cross-region model deployment \citep{liu2023cross,Junyi2024}, this retraining requirement severely constrains their practical applicability. These challenges can be conceptualized as the problem of inductively Forecasting unobserved nodes (iFun), where the key is to develop inductive models that can generalize to unseen sensor configurations without retraining while forecasting states at unobserved locations, as illustrated in \cref{fig:problem}.

The main challenge of iFun lies in modeling heterogeneous spatial relationships without historical data from unobserved sites while adapting to possible changes in sensor deployment configurations. Inductive GNNs, with their message passing mechanism \citep{gilmer2017MPNN}, can generalize to unseen nodes and graph structure \citep{hamilton2017inductive,rossi2018deep}, which aligns well with the requirements of iFun and has achieved initial success in unobserved nodes state inference \citep{Appleby2020KCN,wu2021IGNNK}. However, these studies primarily operate on static graph structures, making it difficult to capture complex spatiotemporal dependencies. On the other hand, while advanced GNNs based on dynamic graph have achieved remarkable success in modeling spatial dependency, they are either transductive architecture with learnable node embeddings \citep{CirsteaSTAware2022ND,liDGCRN2023ND,WengDDGCRN2023ND} or heavily dependent on complete raw node features like dynamic time warping (DTW) \citep{berndt1994DTW,liCIKM2022NS} and graph attention \citep{guo2019ASTGCN,Zheng2020GMAN,wang2023Attention}, making them inapplicable to the iFun problem, which has no historical data on unobserved nodes. These limitations raise a crucial question: how to design an inductive spatiotemporal graph neural network that can effectively capture heterogeneous and dynamic spatial relationships.
Existing research efforts addressing this question have evolved along three distinct directions: (1) augmenting static adjacency matrices through the integration of auxiliary data, such as Points of Interest (POI) and Origin-Destination (OD) data \citep{Zheng2023Increase}, though the accessibility of such auxiliary data remains a practical constraint; (2) attempting to combine static graph structure with learnable node embeddings \citep{WEI2024IAGCN}. However, since these embeddings are only optimized for observed nodes, this approach still heavily relies on static graph structures and achieves only limited improvements with the help of node embedding; and (3) exploring multiple graph message aggregators \citep{wu2021SATC}, drawing inspiration from \cite{Corso2020PNA}. Nevertheless, this approach encounters substantial computational complexity and raises fundamental questions regarding the optimal integration of diverse operators. Our work builds upon and extends the third direction, proposing novel solutions to address its inherent challenges. Beyond these spatial modeling considerations, a temporal dimension further complicates the iFun problem. Existing GNN-based approaches for unobserved node inference primarily address Kriging/state estimation problems \citep{Appleby2020KCN,wu2021IGNNK,Zheng2023Increase,nie2023flowestimation,WEI2024IAGCN,xu2023kits}, which aim to infer current traffic states of unobserved locations using concurrent observations from monitored locations. To date, only a limited number of GNN-based methods \citep{Roth2022Funs,Mei2023UIGNN,sun2023itsc} directly address the challenge of forecasting future states of unobserved nodes, leaving room for a comprehensive solution that addresses both spatial and temporal aspects of the iFun problem.

To address these challenges in both spatial and temporal dimensions of the iFun problem, we propose MoGERNN, a novel inductive spatio-temporal graph representation model. Our approach builds upon two key insights: First, to enhance static graph representation, we design a Mixture of Graph Experts (MoGE) module that integrates multiple specialized message aggregators with a sparse gating mechanism. Unlike traditional static graphs that rely on fixed aggregation patterns, our module captures diverse spatial dependencies through different graph experts, each specializing in distinct neighborhood interaction patterns. The sparse gating network performs two key functions: adaptively routing nodes to their most suitable experts and optimally coupling expert outputs. This design not only achieves more accurate node embeddings but also maintains computational efficiency through selective expert activation. Second, to enable future state prediction beyond current estimation, we develop a Spatio-Temporal Graph-based Encoder-Decoder (STGED) architecture that operates on the powerful node embeddings generated by MoGE, enabling joint spatio-temporal extrapolation and effectively addressing the iFun problem.

The main contributions of this paper are as follows:

\begin{itemize}
    \item Develops a novel method that extends beyond conventional traffic prediction approaches by enabling accurate predictions for unobserved locations without historical data.
    
    \item Designs a sparse-gated Mixture of Graph Experts (MoGE) framework that enhances model expressiveness through multiple specialized graph message aggregators and adaptive expert routing, effectively capturing diverse spatial dependencies in traffic networks.
    
    \item Conducts comprehensive evaluations comparing MoGERNN with various pipeline and end-to-end models for addressing iFun problems from multiple perspectives. Results demonstrate our model's superior performance over state-of-the-art methods in future traffic state prediction on two open-source real-world datasets, with notable improvements for unobserved nodes.
    
    \item Demonstrates robust adaptability to structural changes in sensor networks, maintaining prediction quality during sensor failures and achieving satisfactory zero-shot prediction for newly installed sensors.

\end{itemize}
    
The remainder of the paper is organized as follows: \cref{sec:LR} delivers a literature review related to iFun in traffic data modeling. \cref{sec:pf} provides the formalization of the problem. \cref{sec:mogernn} details the proposed MoGERNN model. \cref{sec:NE} reports the performance and ablation results of the proposed method. \cref{sec:con} concludes the paper.

\section{Literature Review} \label{sec:LR}

\subsection{Traffic Kriging and state estimation}

Both Kriging and TSE aim to estimate the current state of unobserved locations based on limited observations. The most widely used Kriging method is Ordinary Kriging (OKriging) \citep{Wackernagel1995}, which interpolates unseen location by using specific covariance and variogram functions to describe the spatial relationships of random variables at different locations. However, the method assumes the spatial field is stable, which does not correspond to the characteristics of traffic networks. When this stationarity assumption is relaxed, Kriging can be shown to be equivalent to Gaussian Process Regression (GPR), a well-known kernel machine learning method. While GPR offers more flexibility in modeling spatial relationships, it is computationally expensive and thus not applicable to large-scale networks. For large-scale applications, matrix/tensor completion with specific regularization, such as graph Laplacian \citep{Bahadori2014Kriging, nie2023Kriging}, local Markov structure \citep{xiong2010cfiltering}, Gaussian process \citep{Lei2022Kriging}, and temporal consistency \citep{nie2023Kriging}, serves as a effective solution. Notably, \cite{Lei2022Kriging} and \cite{nie2023Kriging} demonstrated satisfactory outcomes in traffic data. Despite this, matrix completion cannot automatically adapt to changes in the sensor network, as it is transductive which requires re-calculation for a new sensor configuration. Moreover, the above Kriging methods cannot be extended to perform temporal forecasting for unobserved nodes, as their underlying theoretical framework is fundamentally designed only for spatial interpolation. 

Recently, inductive methods based on GNNs have been introduced for spatio-temporal Kriging. \cite{Appleby2020KCN} and \cite{wu2021IGNNK} demonstrated the prospect of graph convolution network and diffusion graph convolution network for spatial extrapolation by means of well-designed training procedure. Specifically, 
\citep{wu2021IGNNK} enhanced model generalization to varying sensor configurations through random subgraph construction during the training phase, where different combinations of observed nodes are sampled to simulate potential changes in sensor deployment. However, these two methods are all built on a predefined adjacency matrix, which constrains their ability to capture complex spatial dependencies. To address this challenge, several methods have been proposed. \cite{Zheng2023Increase} incorporated multi-source data such as Points of Interest (POI) and Origin-Destination (OD) information to improve model accuracy. However, such auxiliary data is not always readily available in practical applications. \cite{nie2023flowestimation} proposed a speed pattern-adaptive graph neural network that leverages ubiquitous speed measurements from floating cars and crowd-sourcing vehicles to address traffic volume inference at locations without fixed sensors. However, data integration between government-owned fixed sensors and privately operated mobile sensors faces substantial barriers due to commercial and privacy concerns \citep{wang2024DataFusion}. \cite{Roth2022Funs} introduced Graph Attention mechanism (GAT) that adaptively assigns different weights to neighboring nodes based on their node embeddings. However, in the context of iFun where data for unobserved nodes (virtual sensors) is completely missing, these node-feature driven dynamic neighbor weights can not be learned effectively by attention mechanism, resulting in inaccurate node-wise information aggregation. \cite{WEI2024IAGCN} adapted learnable node embedding-based dynamic graph techniques into an inductive framework and combines them with static graph structure, while introducing an auxiliary task to mitigate information propagation loss caused by random subgraph sampling in IGNNK. However, since these embeddings are optimized through loss functions that require complete node features during training, the resulting dynamic components cannot effectively generalize to unobserved nodes where such features are absent. Inspired by \cite{corso2020principal}, \cite{wu2021SATC} proposed an alternative approach using multiple aggregators for neighbor message passing to capture complex spatial patterns, enhancing the model's generalization capability and accuracy. Nevertheless, this method faces the challenge of effectively integrating multiple aggregators in response to varying graph structures and data contexts. 

Compared to OKriging and tensor completion methods, GNN-based approaches offer the flexibility to map outputs to future timestamps by adjusting training labels, which provides valuable insights for our work. However, it is worth noting that this adaptation is not universally applicable to all GNN-based Kriging methods. For instance, \cite{xu2023kits} identified that traditional GNN-based Kriging methods using decremental training (graph sampling) create a significant graph gap between training and inference. In response, they propose an incremental training strategy by adding fake nodes, but their approach relies fundamentally on a Node-aware Cycle Regularization mechanism that assumes inputs and outputs correspond to the same time interval, rendering these techniques ineffective for future state prediction tasks.

As for TSE, most approaches are based on physical models combined with Kalman Filtering (KF) and its variants \citep{wangRealtimeFreewayTraffic2005, Claudio2016-KF, Sun2018-KF,  WANG2022EKF, MAKRIDIS2023-UKF}. Although filtering methods achieve state-of-the-art (SOTA) performance and are theoretically well-established, their algorithm structure design fundamentally restricts them to one-step-ahead prediction, making them inadequate for advanced traffic control methods \citep{Frejo2012MPC} which requires multi-step traffic state forecasting. Another emerging direction in TSE research is the use of Physics-Informed Neural Network (PINN) \citep{RAISSI2019PINN}. This approach integrates prior traffic knowledge, represented in the form of Partial Differential Equations (PDEs) \citep{lighthill1955kinematic,richards1956shock,payne1971model,whitham1974linear,aw2000resurrection,zhang2002non}, into the neural network framework \citep{Shi2021PIDL,Shi2022PIDL,YUAN2021PIGP,Yuan2022PIGP}, allowing for accurate estimation of states with limited observational data. However, the ability of this method to generalize beyond the spatio-temporal range of the training data is quite limited \citep{Kim2021DPM}. Despite some more advanced studies attempting to address this issue \citep{davini2021piml, fesser2023understanding}, they have not yet been applied to the traffic domain.

\subsection{Traffic forecasting for unobserved locations}
Traffic forecasting has seen rapid development with various sophisticated deep-learning models, including STGNN \citep{yu2018STGCN,li2018dcrnn,Wu2020MTGNN,pems37}, Transformer \citep{DO2019STANN, Liu2023STAEFormer}, CNN \citep{Liu2022SCINet,Wu2023TimesNet}, and MLP-based approaches \citep{Shao2022STID,wang2024timemixer}. However, these methods typically assume the availability of historical observations at all locations and cannot be directly applied to unobserved sites where no historical data exists. For predicting traffic states at unobserved locations, current research efforts primarily follow two technical directions:

The first direction decomposes the problem into independent spatial and temporal extrapolation tasks. \cite{YANG2023EKFLSTM, yang2024EKFTLSTM} proposed an innovative approach that utilizes neural networks to predict future boundary conditions of the road network, which are then fed into the Extended Kalman Filter (EKF) to predict future states of unobserved nodes. However, since Kalman filtering and its variants cannot be integrated into deep learning frameworks for joint training, this decoupled strategy leads to compounded errors from two independent procedures, potentially compromising the overall performance. Similarly, \cite{sun2023itsc} employed Gaussian Mixture Models (GMM) for spatial interpolation of unobserved nodes, followed by a combination of Graph Convolutional Networks and Gated Recurrent Units for global prediction. This approach faces similar limitations as GMM cannot be incorporated into deep learning backpropagation. Moreover, both methods adopt transductive architectures, resulting in limited generalization capabilities.

The second direction embraces end-to-end deep learning approaches, where the entire prediction pipeline can be jointly optimized through backpropagation. \cite{Roth2022Funs} adopted GATv2, an improved variant of Graph Attention Networks (GAT), to model spatiotemporal correlations for global prediction. However, GAT's reliance on complete node features for establishing dynamic spatial relationships significantly constrains its predictive capacity for unobserved nodes and may contaminate the feature representations of observed nodes, thereby degrading overall performance. \cite{Mei2023UIGNN} employed diffusion graph convolution networks for prediction, but this approach, being dependent on static adjacency matrices and lacking comprehensive temporal dynamics modeling, demonstrates limited capability in capturing the complex spatiotemporal characteristics of traffic systems.

\section{Problem Formulation} \label{sec:pf}

In this section, we formally introduce the problem of inductively Forecasting unobserved nodes (iFun). Given a set of \(N\) locations of interest in a traffic network, represented as nodes \(\mathcal{V} = \{\nu_1, \nu_2, \ldots, \nu_N\}\), with \(\widetilde{N}\) nodes (\(\widetilde{N} < N\)) observed by traffic sensors. We model the traffic network as a graph \(\mathcal{G} = (\mathcal{V}, \mathbf{A})\), where \(\mathbf{A}\) denotes the weighted adjacency matrix. Considering the traffic flow characteristics, \(\mathbf{A}\) is defined based on travel distance:
\begin{equation}
\mathbf{A}_{i j}= \begin{cases}\operatorname{exp}\left(-\frac{\operatorname{dist}\left(\nu_i, \nu_j\right)^2}{\sigma^2}\right), & \text { if } \operatorname{dist}\left(\nu_i, \nu_j\right) \leq \kappa \\ 0, & \text { otherwise }\end{cases}
\end{equation}
where $\operatorname{dist}(\nu_i,\nu_j)$ denotes the travel distance from sensor \(\nu_i\) to \(\nu_j\). \(\sigma\) and \(\kappa\) denote the standard deviation of distance samples and threshold, respectively.

The primary target for iFun is to infer the future states of $\overline{N}$ unobserved locations ($\overline{N} = N - \widetilde{N}$) using the historical data from the \(\widetilde{N}\) sensors and graph topology $\mathcal{G}$:
\begin{equation}
    f_{\theta} (\widetilde{\mathbf{X}}^{\widetilde{N} \times P \times I }, \mathcal{G}) = {\mathbf{X}}^{{N} \times F \times O } 
\end{equation}
where $\widetilde{\mathbf{X}}$ and $\overline{\mathbf{X}}$ denote the observed node and unobserved node features, respectively; $P$ and $F$ denote the historical and future time horizon, respectively.

It is essential to highlight that for nodes without sensors, both their number and positions will dynamically change based on the specific requirements during the model's application. Consequently, the complete graph $\mathcal{G}$ is only available during the application phase, also referred to as the test phase in this paper. In addition, considering possible structural changes happens in sensor network, the model must be able to accommodate any addition of new sensors $\widetilde{M}$ and failures of existing sensors $\overline{M}$ in real-time, i.e., no need for model retraining. The mathematical representation is as follows:
\begin{equation}
    f_{\theta} (\widetilde{\mathbf{X}}^{(\widetilde{N}+\widetilde{M}-\overline{M}) \times P }, \mathcal{G}) = \overline{\mathbf{X}}^{ (\overline{N}-\widetilde{M}+\overline{M}) \times F }
\end{equation}

\begin{figure}[t]
    \centering
    \includegraphics[width=1\linewidth]{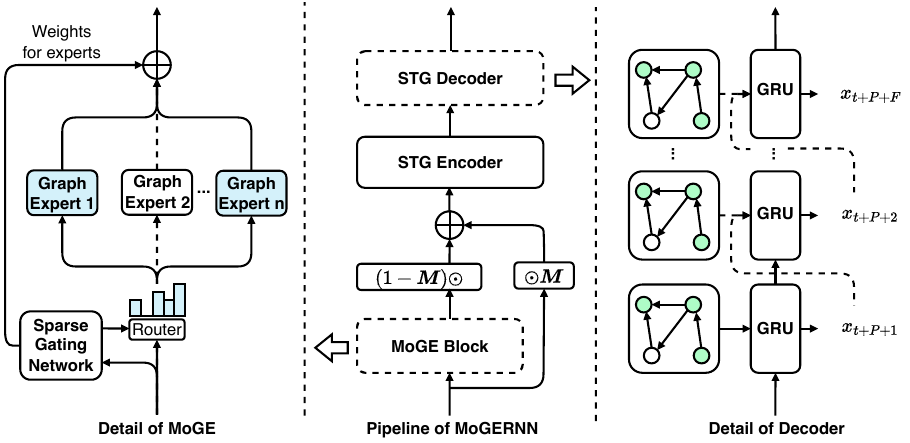}
    \caption{The overview of the proposed MoGERNN.}
    \label{fig:mogernn}
\end{figure}

\section{Methodology} \label{sec:mogernn}

In this section, we introduce the architecture of the proposed MoGERNN. \cref{fig:mogernn} demonstrates the architecture of MoGERNN, it begins with an Mixture of Graph Experts (MoGE) block to model the complex spatio-temporal characteristic, and initially construct the embedding for unobserved nodes. Then, a Spatio-Temporal Graph Encoder-Decoder (STGED) framework is used to obtain multi-step predictions. Details of these two blocks and model training and testing are introduced in the following subsection.

\subsection{Mixture of Graph Experts For Unobserved Node Embedding}

To model the heterogeneous and complex spatio-temporal characteristic, the following aggregators that are commonly used in spatio-temporal applications are introduced \citep{gilmer2017neural, kipf2017semisupervised,xu2018how,corso2020principal}.

\begin{itemize}
    \item Distance-based weighted aggregator: $\mathbf{X}_j^{(l+1)}=\frac{\sum_{i \in \mathcal{N}(j)} \mathbf{X}_i^{(l)} \cdot \mathbf{A}_{i j}} {\sum_{i \in \mathcal{N}(j)} \mathbf{A}_{i j}}$
    \item Mean aggregator: $ \mathbf{X}_j^{(l+1)} = \operatorname{mean}_{i \in \mathcal{N}(j)} \mathbf{X}_i^{(l)}$
    \item Max pooling aggregator: $\mathbf{X}_j^{(l+1)} = \max_{i \in \mathcal{N}(j)} \mathbf{X}_i^{(l)}$
    \item Min pooling aggregator: $\mathbf{X}_j^{(l+1)} = \min_{i \in \mathcal{N}(j)} \mathbf{X}_i^{(l)}$
    \item Diffusion convolution aggregator:
    
    $\mathbf{X}_j^{(l+1)} =\sum_{k=0}^{K-1} (\sum_{i}(\mathcal{D}_O^{-1} \mathbf{A})^k_{ij}\mathbf{X}_i^{(l)}\mathbf{W}_{k,O}^{(l)} +\sum_{i}(\mathcal{D}_I^{-1} \mathbf{A}^\top)^k_{ij}\mathbf{X}_i^{(l)}\mathbf{W}_{k,I}^{(l)})$
\end{itemize}
where $\mathbf{X}^{(l)} \in \mathbb{R}^{N \times H_1}$ denotes the input graph signal at layer $l$. \( \mathcal{N}(j) \) denotes the set of all nodes \( i \) that are neighbors of node \( j \), i.e., all nodes \( i \) such that \( \mathbf{A}_{ij} > 0 \). $\mathcal{D}_O$ and $\mathcal{D}_I$ denote out-degree and in-degree of adjacency matrix $\mathbf{A}$, respectively. $\mathbf{W}_{k,O}^{(l)}$ and $\mathbf{W}_{k,I}^{(l)}$ represent learnable parameters in $\mathbb{R}^{H_1\times H_1}$ at step $k$, layer $l$, respectively. In the following text, the graph aggregation operation is denoted as  \(\operatorname{Agg}\).

The selection of these aggregators is motivated by both theoretical foundations and empirical evidence in traffic modeling. The distance-weighted and mean aggregators serve complementary purposes: the former aligns with Tobler's First Law of Geography \citep{tobler1970firstlaw} by assigning higher weights to spatially closer nodes, while the latter treats all neighbors equally to capture potential long-range dependencies. The max-pooling and min-pooling aggregators are specifically designed to capture distinct traffic states - max-pooling for free-flow conditions characterized by higher speeds, and min-pooling for congested states marked by lower speeds. Additionally, the diffusion convolution aggregator is included based on its proven effectiveness in modeling traffic flow dynamics \citep{li2018dcrnn}.

In order to avoid the effective of unobserved nodes to observed ones and further direct the network to focus on acquiring information from neighboring nodes, we make the following adjustments to the adjacency matrix used in the aggregators within first layer:

\begin{equation}
    \mathbf{A}'_{i j} = \begin{cases} 0, & \text {if} \quad i==j, \text{ or node $j$ is not observed}   \\ \mathbf{A}_{i j}, & \text { otherwise }\end{cases}
     \label{eq:no-self}
\end{equation}

Inspired by the Mixture of Experts (MoE) proposed in Large Language Model (LLM), we design a Mixture of Graph Experts (MoGE)  to efficiently integrate these aggregators. MoGE assigns weights to different experts through a sparse gating network, which reduces the complexity of the model by activating only some of the experts at a time compared to dense gating. Existing research has shown that the MoE architecture not only enhances the model capacity without adding additional computational resources, but also integrates the strengths of different experts to improve the performance of the overall task \citep{Li2023STMoE,lee2024testam}.

\textbf{Graph Experts}: Multiple graph experts are constructed based on different types of aggregators. Specifically, in each graph expert, the input features $\mathbf{X} \in \mathbb{R}^{N \times P}$ are first embedded into the latent space $\mathbb{R}^{N\times H_2}$ with a learnable parameter $W_1 \in \mathbb{R}^{P \times H_2}$, then the features from different locations are mixed using the graph aggregator $\operatorname{Agg}$ defined above, and finally mapping the features back to the space $\mathbb{R}^{N\times P}$ through a linear layer and activation function $\operatorname{Act}$:
\begin{equation}
    \mathbf{X}_e = \operatorname{Act}(\operatorname{Linear}(\operatorname{Agg}_e(\mathbf{X}W_1,\mathbf{A})))
\end{equation}
where $\mathbf{X}_e$ denotes the output of $e$-th graph expert based on $e$-type graph aggregator.

\textbf{Sparse Gating Network}: To obtain scores \(S\) for each expert, input features \(\mathbf{X}\) that are fed into the graph experts are also processed through a scoring network composed of two linear layers and an activation function:
\begin{equation}
    \mathbf{S} =\operatorname{Linear}(\operatorname{Act}(\operatorname{Linear}(\mathbf{X})))
\end{equation}
where $\mathbf{S} \in \mathbb{R}^{N \times N_e}$, and \(N_e\) denotes the number of graph experts.

Softmax function is used to convert the scores into weights $\mathbf{w}$ for different experts. To reduce the complexity, only top $K$ experts are activated:
\begin{equation}
    \mathbf{w} = \operatorname{Softmax}(\operatorname{KeepTopK}(\mathbf{S},K)) 
    \label{eq:softmax}
\end{equation}

\begin{equation}
    \operatorname{KeepTopK}(\mathbf{S}, K) = \begin{cases} 
    \mathbf{S}_e, & \text{if } \mathbf{S}_e \text{ is among the top } K \text{ elements} \\ 
    -\infty, & \text{otherwise}
    \end{cases}
    \label{eq:keepk}
\end{equation}

Based on \cref{eq:softmax} and \cref{eq:keepk}, the sparse gating network can zero out certain expert weights, enabling selective expert activation during inference. As shown in \cref{fig:mogernn}, dashed lines represent inactive pathways. The sparse gating network plays a crucial role in our model by serving multiple purposes. Theoretically, it creates an information bottleneck that serves as implicit regularization, which is particularly important as the combination of different graph neural experts results in a complex model prone to overfitting. From a practical perspective, by activating only a subset of experts, it significantly reduces computational overhead while maintaining model effectiveness through optimal expert selection and weight assignment.

Given graph experts and sparse gating network, the output of MoGE will be $\mathbf{X}_{\text{moge}} = \sum_e \mathbf{w}_e \odot \mathbf{X}_e$. In this work, MoGE is designed specifically for unobserved node embedding construction, the states of observed nodes remain unchanged before entering the encoder-decoder.:
\begin{equation}
    \mathbf{X} = \mathbf{X} \odot \mathbf{M} + \mathbf{X}_{\text{moge}} \odot (1-\mathbf{M})
\end{equation}
where $\mathbf{M} \in \mathbb{R}^{N \times P}$ is a binary matrix, with zeros indicating unobserved entities. Moreover, $\mathbf{X} \in \mathbb{R}^{N \times P}$ is reorganized as $\mathbf{X} \in \mathbb{R}^{N \times P \times 1}$, for the sake of operation in encoder-decoder.

\subsection{Graph Encoder-Decoder For Spatio-Temporal Extrapolation}
An encoder-decoder incorporating Recurrent Neural Networks (RNNs) is a widely used multi-step prediction framework that has achieved satisfactory results in traffic prediction \citep{Wang2021ED,Kong2024ED}. In this paper, we use the GRU-based Encoder-Decoder framework. Compared to standard RNNs and Long-Short Term Memory (LSTM), GRU requires less memory and fewer parameters while maintaining comparable performance, and has been demonstrated to couple well with GNN \citep{li2018dcrnn,cini2022filling}. Inspired by \cite{li2018dcrnn}, we replace the linear transformation in GRU with graph aggregators:

\begin{equation}
    \mathbf{r}_t = \operatorname{sigmoid}\left(\operatorname{Agg}_r\left(\mathbf{X}_t||\mathbf{h}_{t-1}\right) + \mathbf{b}_r\right)
\end{equation}
\begin{equation}
    \mathbf{u}_t = \operatorname{sigmoid}(\operatorname{Agg}_u(\mathbf{X}_t||\mathbf{h}_{t-1})+\mathbf{b}_u)
\end{equation}
\begin{equation}
    \mathbf{c}_t = \operatorname{tanh}(\operatorname{Agg}_c(\mathbf{X}_t|| (\mathbf{r}_t \odot \mathbf{h}_{t-1})) + \mathbf{b}_c)
\end{equation}    
\begin{equation}
    \mathbf{h}_t = \mathbf{u}_t \odot \mathbf{h}_{t-1} + (1-\mathbf{u}_t) \odot \mathbf{c}_t
\end{equation}
where $||$ denotes the concatenation operator, $\mathbf{h}_t \in \mathbb{R}^{N \times H_2}$ is the hidden value that contains compressed information about the sequence before time instance $t$. In our design, $\operatorname{Agg}_r$, $\operatorname{Agg}_u$, and $\operatorname{Agg}_c$, which represent the graph aggregators for the reset gate, update gate, and candidate hidden state respectively, share the same type of graph aggregator.

Two GRUs are initialised to act as encoder and decoder respectively. In the encoder, the historical sequence $\mathbf{X}_t,\mathbf{X}_{t+1},...,\mathbf{X}_{t+P}$ is compressed to a hidden variable $\mathbf{h}_{t+P}$. This variable is used to initialize the hidden variable in the decoder. Different from encoder, decoder additionally contains a linear output layer, where the output of each step of decoder will be used as input for the next step.

\begin{algorithm}[t]
\caption{Inductive Training Procedure}
\label{alg:train}
\begin{algorithmic}[1]
\Require Data $\widetilde{\mathbf{X}} \in \mathbb{R}^{(\widetilde{N}, L)}$, graph $\widetilde{\mathcal{G}}$, historical length $P$, prediction length $F$, stride $s$, batch size $B$, epochs $E$, learning rate $\eta$, mask rate $m$, maximum epoch for training $E_m$.
\Ensure Model parameters $\theta$
\State Create training pairs $(\widetilde{\mathbf{X}}_{\text{train}}, \widetilde{\mathbf{y}}_{\text{train}})$ using sliding window with stride $s$
\State Initialize $\theta$
\For{epoch $= 1$ to $E$}
    \For{each mini-batch $(\widetilde{\mathbf{X}}_b, \widetilde{\mathbf{y}}_b)$ after random shuffle}
        \State Mask random nodes in $\widetilde{\mathbf{X}}_b$ with rate $m$
        \State $r = \max(1-\text{epoch}/E_m,0)$ \Comment{Teacher forcing rate}
        \State $\mathbf{y}^- = f_{\theta}(\widetilde{\mathbf{X}}_b,\widetilde{G},r,\widetilde{\mathbf{y}}_b)$
        \State Update $\theta$ using $\nabla_{\theta}\mathcal{L}(\mathbf{y}^-, \widetilde{\mathbf{y}}_b)$
    \EndFor
\EndFor
\State \Return $\theta$
\end{algorithmic}
\end{algorithm}

\subsection{Model Training and Testing} \label{sec:train}

To effectively forecast the unobserved nodes, an inductive training algorithm is developed. This method leverages random masking to simulate the test environment based on observed training data. By creating diverse input-output pairs, this approach enhances the model's robustness and ensures it can generalize to new, unseen nodes during application. 

Suppose we have collected training data $\widetilde{\mathbf{X}} \in \mathbb{R}^{(\widetilde{N}, L)}$ from \(\widetilde{N}\) observed locations over a time horizon of length \(L\). The sensor network can be represented as a graph \(\widetilde{\mathcal{G}} = (\widetilde{\mathcal{V}}, \widetilde{\mathbf{A}})\), where \(\widetilde{\mathcal{V}} = \{\widetilde{\nu}_1, \widetilde{\nu}_2, \dots, \widetilde{\nu}_{\widetilde{N}}\}\) denotes the set of nodes corresponding to the sensors. The inductive training algorithm is detailed in \cref{alg:train}. The core distinction of our training algorithm compared to conventional traffic prediction training algorithms lies in the use of random masking for each batch of data (lines 9-11 in \cref{alg:train}). Inspired by \citep{wu2021IGNNK}, we randomly select a subset of nodes and set their features to zero in each batch. This operation not only simulates the data conditions during model application for predicting unobserved nodes, but also, by masking different nodes in different batches, replicates sensor failures and additions, forming the cornerstone of the model's successful application to iFun. In addition, to enhance the model training, scheduled sampling \citep{bengio2015scheduled} with linear decay strategy is employed. Specifically, we initially feed the true observation sequence within the prediction horizon into the decoder, i.e., teacher forcing, with a high probability, then decrease this probability linearly, and finally switch to using the decoder's prediction as the next step input, as shown in the third panel of \cref{fig:mogernn}.



During inference, random masking is no longer required. For unobserved nodes of interest, we assume the presence of virtual sensors \citep{VirtualSensor2009,VirtualSensor2021}. The graph $\widetilde{G}$ is replaced by the complete graph $G$, which includes both real and virtual sensors. The input $\widetilde{\mathbf{X}}$ is replaced by the features of the currently observed and unobserved nodes, denoted as $\mathbf{X}$, with the features of the unobserved nodes set to zero. The mathematical formulation for the inference is as follows:

\begin{equation}
    \mathbf{y}^- = f_{\theta ^*}(\mathbf{X},\mathcal{G})
    \label{eq:inference}
\end{equation}
where $\theta^*$ represents the parameters of the trained model.

\begin{table}[t]
  \centering
   \caption{Summary of datasets}
   \resizebox{\columnwidth}{!}{%
   \begin{tabular}{cccccc}
    \toprule
    \textbf{Dataset} & \textbf{Unit} & \textbf{Sensors Number}  & \textbf{Time Period} & \textbf{Sampling Frequency} & \textbf{Missing Rate} \\
    \hline
    METR-LA & mph & 207  & 01/03/2012 - 30/06/2012 & 5 min &  8.11\%\\
    PEMS-BAY & mph & 325  & 01/01/2017 - 31/05/2017 & 5 min & 0.00\%\\
    \bottomrule
   \end{tabular}}
  \label{tab:dataset}
\end{table}

\section{Numerical Experiments} \label{sec:NE}

In this section, several experiments are set to answer the following question:

\begin{enumerate}[label=Q\arabic*:,itemsep=0pt, parsep=0pt]
    \item \textbf{Prediction performance} (\cref{sec:q1}). Does the prediction performance of MoGERNN outperform other benchmark methods considering unobserved nodes?
    \item \textbf{Adaptability to the structural changes of sensing networks} (\cref{sec:q2}). Is MoGERNN capable of adapting to the possible structural changes in sensor networks, and how does its performance differ from the re-trained model?
    \item \textbf{Impact of the VS-to-AAS ratio} (\cref{sec:q3}). How does the prediction performance of MoGERNN at unobserved locations vary with the ratio of virtual to available sensors?
    \item \textbf{Ablation study} (\cref{sec:q4}). Are MoGE and Encoder-Decoder components effective?
    \item \textbf{Parameters study} (\cref{sec:q5}. How the number of activated experts affects performance and computation time?
\end{enumerate}

\subsection{Experimental Setup}
\noindent \textbf{Dataset.} Two widely used open-source traffic speed datasets are evaluated, including METR-LA and PEMS-BAY from \cite{li2018dcrnn}. Descriptions for the datasets are as shown in \cref{tab:dataset}.

\vspace{0.2cm}
\noindent \textbf{Baselines.} We selected eight methods as benchmarks. These baselines can be categorized into four groups: the combination of Kriging and local traffic prediction methods (OKriging+ED, KNN+ED), state-of-the-art local traffic prediction STGNNs adapted for the iFun scenario (TGCN, GMSDR), modified inductive GNN-based Kriging approaches (IGNNK, INCREASE), and methods specifically designed for the iFun problem (FUNS, UIGNN). The detailed is as follows:

\begin{itemize}
    \item[(1)] OKriging\footnote{\url{https://github.com/GeoStat-Framework/PyKrige}}\citep{saito2005OKriging} with ED:  This method combines Ordinary Kriging with encoder-decoder architecture in MoGERNN. The approach first employs Ordinary Kriging to interpolate values at unobserved locations using Gaussian variogram functions to model spatial relationships. After training ED inductively, we feed both the Kriging-interpolated values and observed measurements into ED to predict future states for both observed and unobserved nodes. 
    \item[(2)] KNN\footnote{\url{https://scikit-learn.org/stable/modules/generated/sklearn.impute.KNNImputer.html}} with ED: Similar to the first baseline, OKriging is replace by K-nearest neighbors. KNN performs interpolation based on the average of k-nearest neighbors. Neighbors are defined as different samples that exhibit "feature similarity." Since the speed feature of unobserved nodes are completely missing, we utilize auxiliary features such as latitude, longitude, and timestamps to facilitate interpolation for these unobserved nodes. 
    \item[(3)] TGCN\footnote{\url{https://github.com/lehaifeng/T-GCN/tree/master/T-GCN}} \citep{Zhao2020TGCN}: TGCN integrates graph convolution network and temporal convolution network to model the spatio-temporal dynamics of traffic. As it adopts static-graph, we can modified it towards iFun scenario by using the proposed inductive training procedure.
    \item[(4)] GMSDR\footnote{\url{https://github.com/dcliu99/MSDR/tree/master/GMSDR_Flow/model/pytorch}} \citep{liu2022GMSDR}: GMSDR is a spatial-temporal forecasting framework that captures long-range dependencies by explicitly incorporating multiple historical states in its graph-based architecture. Since GMSDR employs node embeddings as part of its architecture, we train it in a transductive manner to learn node-specific representations.
    \item[(5)] IGNNK\footnote{\url{https://github.com/Kaimaoge/IGNNK}} \citep{wu2021IGNNK}: IGNNK develops an inductive training procedure to estimate the unobserved locations state, the main block is consisted by three diffusion graph convolution layers using bidirectional random walk diffusion kernel. Instead of training the model to perform traditional Kriging interpolation, we modify IGNNK learning objective to directly predict future states at both observed and unobserved nodes, making it compatible with the iFun problem setting.
    \item[(6)] INCREASE{\footnote{\url{https://github.com/zhengchuanpan/INCREASE}}} \citep{Zheng2023Increase} : Inductive graph representation learning model for spatio-temporal Kriging, featuring GNN-based spatial aggregation and GRU-based temporal modeling to capture heterogeneous spatial relations with various temporal patterns for interpolating state from unseen nodes. We modify it as same as baseline (5).
    \item[(7)] FUNS\footnote{\url{https://github.com/roth-andreas/funs}} \citep{Roth2022Funs}, an inductive spatio-temporal graph neural network (STGNN) that combines a Message Passing Neural Network (MPNN) \citep{gilmer2017MPNN} framework with a Gated Recurrent Unit (GRU), where the message aggregation function employs the variant of graph attention network (GAT), GATv2 \citep{brody2022GATv2}.
    \item[(8)] UIGNN\footnote{\url{https://github.com/lijunxian111/UIGNN}} \citep{Mei2023UIGNN}, an uncertainty-aware inductive graph neural network framework that utilizes Diffusion Graph Convolution Networks (DGCN), which is an extension method of IGNNK proposed by \cite{wu2021IGNNK}. To ensure a fair comparison, the independent uncertainty quantification module in UIGNN is toggled off.
\end{itemize}

\vspace{0.2cm}
\noindent \textbf{Metrics.} All the methods are evaluated by three commonly adopted metrics in time series prediction: Mean Absolute Percentage Error (MAPE), Mean Absolute Error (MAE) and Root Mean Squared Error (RMSE). It is important to note that original missing values in the dataset are excluded from the evaluation.

\vspace{0.2cm}
\noindent \textbf{Implementation.} The proposed MoGERNN is implemented with PyTorch 2.2.1 on an NVIDIA RTX 4090 GPU. The historical $P$ and prediction horizons $F$ are both fixed in 12 steps (1 hour). Mean square error is adopted as the loss function. The dataset is divided into training and testing sets in a 7:3 ratio. The validation set is kept the same as the training set and used for early stopping with patience set to 10. We collect data using a sliding window with a stride of 12. In the training procedure, the random mask rate is set as 25\%. Following \citep{li2018dcrnn}, the diffusion convolution aggregator is chosen to be the $\operatorname{Agg}$ in Encoder-Decoder. To validate model performance for the iFun problem, different ratios of Virtual Sensors (VS), Failed Sensors (FS), Newly Added Sensors (NAS) and Always Available Sensors (AAS) are set. For clarity, we only give the formal definition of each type of sensors here, the ratio of them will be illustrated in each case study. The code for MoGERNN implementation is publicly available at: \url{https://github.com/ZJU-TSELab/MoGERNN}.

\begin{itemize}
    \item VS: Sensors installed in never unobserved nodes and used solely for evaluation during the test phase.
    \item FS: Sensors that function normally during the training phase but are excluded in input during the testing phase.
    \item NAS: Sensors that are newly added in the test phase.
    \item AAS: Sensors that are always available during train and test.
\end{itemize}

\begin{table}[t]
\caption{Performance at three prediction horizons on METR-LA and PEMS-BAY. The best result is highlighted in bold, while the second-best result is underlined. The "Improve" illustrates the percentage of relative performance improvement of MOGERNN compared to the most competitive baseline.}
\label{tab:PP}
\vspace{-0.7em}
\resizebox{\textwidth}{!}{%
\begin{tabular}{cllccccccccc}
\toprule
\multirow{2}{*}{\textbf{Dataset}}   & \multirow{2}{*}{\textbf{Sensor}} & \multirow{2}{*}{\textbf{Model}} & \multicolumn{3}{c}{\textbf{15 min}} & \multicolumn{3}{c}{\textbf{30 min}} & \multicolumn{3}{c}{\textbf{60 min}} \\ \cline{4-12}
                           &                         &                        & \textbf{MAPE}    & \textbf{MAE}    & \textbf{RMSE}    & \textbf{MAPE}    & \textbf{MAE}    & \textbf{RMSE}    & \textbf{MAPE}    & \textbf{MAE}    & \textbf{RMSE}    \\
\hline
\multirow{20}{*}{\rotatebox{90}{METR-LA}}  

                                           & \multirow{10}{*}{Virtual sensor}     & {OKriging+ED}            & 21.71   & 8.22   & 11.21   & 22.33   & 8.43   & 11.47   & 22.68   & 8.83   & 11.91    \\
                                           &                         & {KNN+ED}                 & 21.05   & 7.97   & 13.24   & 20.72   & 7.91   & 12.84   & 20.06   & 7.88   & 12.49    \\
                                           &                         & {\color{black}{GMSDR}}     & {\color{black}{25.10}} & {\color{black}{9.25}} & {\color{black}{11.95}} & {\color{black}{25.49}} & {\color{black}{9.49}} & {\color{black}{12.15}} & {\color{black}{24.72}} & {\color{black}{9.54}} & {\color{black}{12.18}}      \\
                                           & & {\color{black}{TGCN}} & {\color{black}{21.43}} & {\color{black}{7.60}} & {\color{black}{10.76}} & {\color{black}{22.17}} & {\color{black}{7.82}} & {\color{black}{11.05}} & {\color{black}{22.43}} & {\color{black}{8.06}} & {\color{black}{11.32}}     \\
                                           & & {\color{black}{IGNNK}} & {\color{black}{17.58}} & {\color{black}{6.64}} & {\color{black}{\ul9.82}} & {\color{black}{18.57}} & {\color{black}{\ul6.90}} & {\color{black}{\ul10.21}} & {\color{black}{19.41}} & {\color{black}{\ul7.31}} & {\color{black}{\ul10.66}}     \\
                                           &                         & {\color{black}{INCREASE}} & {\color{black}{20.20}}    & {\color{black}{8.50}}   & {\color{black}{12.30}}    & {\color{black}{19.91}}    & {\color{black}{8.43}}   & {\color{black}{12.31}}    & {\color{black}{19.15}}    & {\color{black}{8.26}}   & {\color{black}{12.19}} \\
                                           &                         & {UIGNN}                  & {\ul16.70}   & {\ul6.62}   & 9.89    & {\ul17.88}   & 6.93   & 10.29   & {\ul18.92}   & 7.33   & 10.76    \\
                                           &                         & {FUNS}                   & 21.64   & 8.12   & 11.37   & 21.50    & 8.04   & 11.30    & 20.51   & 7.74   & 10.93    \\
                                           &                         & {MoGERNN (ours)}         & \textbf{14.08}   & \textbf{5.83}   & \textbf{8.64}    & \textbf{15.06}   & \textbf{5.97}   & \textbf{8.88}    & \textbf{15.91}   & \textbf{6.19}   & \textbf{9.38}     \\ \cline{3-12} 
\multicolumn{1}{l}{}                       &                          & Improve (\%)             & +15.69            & +11.93           & +12.02   & +15.77  & +13.48            & +13.03            & +15.91        & +15.32   & +12.01       \\\cline{2-12}

                                           & \multirow{10}{*}{Always available sensor}    & {OKriging+ED}            & 8.38    & 3.10   & 5.37    & 11.44   & 3.86   & 6.75    & 14.56   & 4.85   & 8.31    \\
                                           &                         & {KNN+ED}                 & 8.30    & {\ul3.09}   & {\ul5.35}    & {\ul10.89}   & {\ul3.84}   & {\ul6.54}    & {\ul13.11}   & {\ul4.52}   & {\ul7.87}    \\
                                           &                         & {\color{black}{GMSDR}}    & {\color{black}{9.44}}    & {\color{black}{3.55}}   & {\color{black}{6.09}}    & {\color{black}{11.14}}   & {\color{black}{4.01}}   & {\color{black}{6.90}}    & {\color{black}{13.86}}   & {\color{black}{4.78}}   & {\color{black}{8.00}}   \\
                                           &                         & {\color{black}{TGCN}}    & {\color{black}{13.81}}    & {\color{black}{5.32}}   & {\color{black}{7.74}}    & {\color{black}{15.94}}   & {\color{black}{5.84}}   & {\color{black}{8.66}}    & {\color{black}{18.48}}   & {\color{black}{6.61}}   & {\color{black}{9.85}}   \\
                                           &                         & {\color{black}{IGNNK}}    & {\color{black}{8.50}}    & {\color{black}{3.34}}   & {\color{black}{5.76}}    & {\color{black}{10.96}}   & {\color{black}{3.98}}   & {\color{black}{7.06}}    & {\color{black}{14.11}}   & {\color{black}{4.95}}   & {\color{black}{8.58}}   \\
                                           &                         & {\color{black}{INCREASE}} & {\color{black}{-}}      & {\color{black}{-}}    & {\color{black}{-}}   & {\color{black}{-}}    & {\color{black}{-}}   & {\color{black}{-}}    & {\color{black}{-}}   & {\color{black}{-}}   & {\color{black}{-}} \\
                                           &                         & {UIGNN}                  & {\ul8.42}    & 3.31   & 5.72    & 10.91   & 3.95   & 6.95    & 13.98   & 4.85   & 8.45    \\
                                           &                         & {FUNS}                   & 16.27   & 5.66   & 9.11    & 16.52   & 5.73   & 9.18    & 15.84   & 5.53   & 8.88    \\
                                           &                         & {MoGERNN (ours)}         & \textbf{8.07}    & \textbf{3.04}   & \textbf{5.31}    & \textbf{10.45}   & \textbf{3.61}   & \textbf{6.45}    & \textbf{12.74}   & \textbf{4.24}   & \textbf{7.66}    \\ \cline{3-12} 
\multicolumn{1}{l}{}                        &                                  & Improve (\%)                    & +2.77                 & +1.62                 & +0.75                 & +4.04                 & +5.99                 & +0.16                 & +2.82                 & +6.19                 & +2.67                 \\

\hline
\multirow{20}{*}{\rotatebox{90}{PEMS-BAY}} 
                                           & \multirow{10}{*}{Virtual sensor}     & {OKriging+ED}            & 11.19   & 4.83   & 7.86    & 11.41   & 4.88   & 8.03    & 11.55   & 4.98   & 8.39      \\
                                           &                         & {KNN+ED}                 & 13.91   & 5.89   & 11.04   & 13.53   & 5.55   & 10.33   & 12.95   & 5.18   & 9.67      \\
                                           &                         & {\color{black}{GMSDR}}    & {\color{black}{12.82}}    & {\color{black}{5.67}}   & {\color{black}{8.27}}    & {\color{black}{13.11}}    & {\color{black}{5.80}}   & {\color{black}{8.45}}    & {\color{black}{12.54}}    & {\color{black}{5.49}}   & {\color{black}{8.31}}     \\
                                           &                         & {\color{black}{TGCN}}    & {\color{black}{9.51}}    & {\color{black}{3.94}}   & {\color{black}{6.78}}    & {\color{black}{10.12}}    & {\color{black}{4.09}}   & {\color{black}{7.11}}    & {\color{black}{10.87}}    & {\color{black}{4.36}}   & {\color{black}{7.61}}     \\
                                           &                         & {\color{black}{IGNNK}}    & {\color{black}{\ul8.05}}    & {\color{black}{\ul3.59}}   & {\color{black}{6.12}}    & {\color{black}{\ul8.77}}    & {\color{black}{\ul3.81}}   & {\color{black}{\ul6.29}}    & {\color{black}{9.78}}    & {\color{black}{4.18}}   & {\color{black}{6.98}}     \\
                                           &                         & {\color{black}{INCREASE}}    & {\color{black}{9.46}}    & {\color{black}{4.02}}   & {\color{black}{6.85}}    & {\color{black}{9.53}}    & {\color{black}{4.04}}   & {\color{black}{6.87}}    & {\color{black}{\ul9.57}}    & {\color{black}{\ul4.13}}   & {\color{black}{6.97}}     \\
                                           &                         & {UIGNN}                  & 8.24    & 3.67   & \textbf{5.92}    & 8.92    & 3.88   & 6.31    & 9.83    & 4.22   & {\ul6.94}      \\
                                           &                         & {FUNS}                   & 10.98   & 4.72   & 7.67    & 10.90   & 4.75   & 7.65    & 10.75   & 4.72   & 7.69      \\
                                           &                         & {MoGERNN (ours)}         & \textbf{7.58}    & \textbf{3.56}   & {\ul5.95}    & \textbf{8.10 }   & \textbf{3.68}   & \textbf{6.20}    & \textbf{8.88}    & \textbf{3.92}   & \textbf{6.72}      \\  \cline{3-12} 
\multicolumn{1}{l}{}                       &                         & Improve (\%)             & +5.84                 & +0.84                 & -0.51                & +7.64                 & +3.41                 & +1.53                 & +7.21                 & +5.08                 & +3.17                 \\\cline{2-12}
                                           & \multirow{10}{*}{Always available sensor}    & {OKriging+ED}            & 3.43    & {\ul1.57}   & 3.13    & 5.10    & {\ul2.06}   & 4.62    & 6.99    & {\ul2.67}   & 6.08      \\
                                           &                         & {KNN+ED}                 & 3.52    & 1.61   & 3.18    & 5.21    & 2.11   & 4.73    & 7.28    & 2.75   & 6.23      \\
                                           &                         & {\color{black}{GMSDR}}    & {\color{black}{4.86}}    & {\color{black}{2.18}}   & {\color{black}{4.10}}    & {\color{black}{5.59}}    & {\color{black}{2.46}}   & {\color{black}{4.66}}    & {\color{black}{\ul 6.88}}    & {\color{black}{2.89}}   & {\color{black}{\ul5.47}}    \\
                                           &                         & {\color{black}{TGCN}}    & {\color{black}{6.50}}    & {\color{black}{2.91}}   & {\color{black}{4.91}}    & {\color{black}{7.62}}    & {\color{black}{3.24}}   & {\color{black}{5.69}}    & {\color{black}{9.50}}    & {\color{black}{3.83}}   & {\color{black}{6.91}}    \\
                                           &                         & {\color{black}{IGNNK}}    & {\color{black}{3.40}}    & {\color{black}{1.63}}   & {\color{black}{3.12}}    & {\color{black}{5.11}}    & {\color{black}{2.19}}   & {\color{black}{4.44}}    & {\color{black}{7.52}}    & {\color{black}{3.02}}   & {\color{black}{6.00}}    \\
                                           &                         & {\color{black}{INCREASE}} & {\color{black}{-}}      & {\color{black}{-}}    & {\color{black}{-}}   & {\color{black}{-}}    & {\color{black}{-}}   & {\color{black}{-}}    & {\color{black}{-}}   & {\color{black}{-}}   & {\color{black}{-}}    \\
                                           &                         & {UIGNN}                  & {\ul3.37}    & 1.63   & {\ul3.06}    & {\ul4.92}    & 2.16   & {\ul4.34}    & 7.09    & 2.94   & 5.83      \\
                                           &                         & {FUNS}                   & 8.06    & 3.24   & 6.37    & 7.82    & 3.21   & 6.30    & 7.56    & 3.11   & 6.30      \\
                                           &                         & {MoGERNN (ours)}         & \textbf{3.20}    & \textbf{1.54}   & \textbf{2.93}    & \textbf{4.55}    & \textbf{1.96}   & \textbf{4.06}    & \textbf{6.22}    & \textbf{2.50}   & \textbf{5.29}      \\ \cline{3-12} 
\multicolumn{1}{l}{}                        &                                  & Improve (\%)                    & +5.04                 & +1.91                 & +4.25                 & +7.52                 & +4.85                 & +6.45                 & +9.59                 & +6.37                 & +3.29                 \\
\bottomrule
\end{tabular}}
\end{table}

\begin{figure}[]
    \centering
    \includegraphics[width=0.9\linewidth]{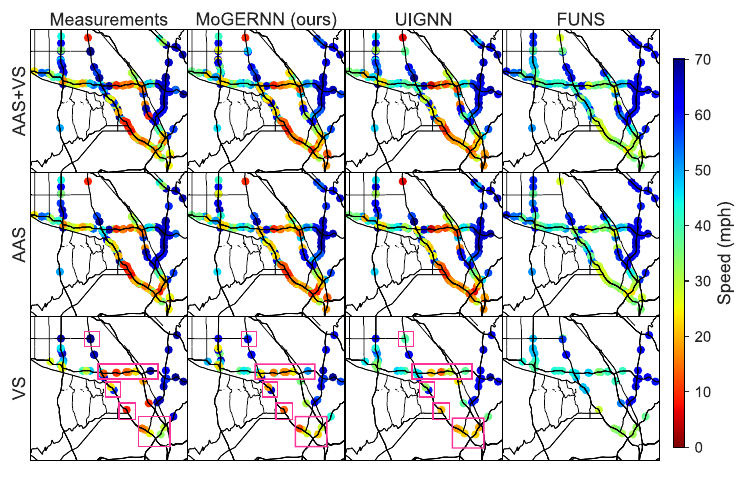}
    \vspace{-1em}
    \caption{Map presentation of prediction performance in an evening-peak time point (2012-05-24 17:30) of METR-LA.}
    \label{fig:map}
\end{figure}

\subsection{Prediction Performance on Unobserved and Observed Locations (Q1)} \label{sec:q1}

In this experiment, 25\% sensors are excluded randomly during training so as to play a VS role in the test, with the remaining sensors serving as AAS. \cref{tab:PP} summarizes the performance over 15, 30, and 60-minute prediction horizons. While our study focuses more on the predictions for VS, we also present the performance of AAS to comprehensively show the performance of the proposed method. As shown in \cref{tab:PP}, our MoGERNN outperforms baseline models across all prediction horizons on both datasets for unobserved locations, with particularly significant improvements on METR-LA. Additionally, our model maintains comprehensive advantages in most cases for future state prediction at observed stations. 

Among the baseline methods, the concatenated approaches combining kriging with deep learning models (OKriging+ED, KNN+ED) show inferior performance. While previous research has shown that KNN and OKriging perform competitively in spatial extrapolation tasks \citep{wu2021IGNNK}, these concatenated approaches struggle with spatiotemporal extrapolation due to error propagation between multiple independent models. For conventional traffic prediction models, despite GMSDR's state-of-the-art performance in standard traffic prediction through dynamic graph construction, its performance deteriorates significantly in the iFun task, as it fails when unobserved nodes lack historical data for node embedding training. TGCN, while using a static graph structure, performs better than GMSDR at unobserved locations due to its preserved spatial connectivity information, though its overall performance is limited by the over-smoothing issue. In terms of methods extending inductively kriging approaches, IGNNK shows strong generalization capability to unseen nodes through its diverse subgraph construction strategy. However, its performance is constrained by inadequate temporal dynamics modeling. INCREASE, despite considering both spatial and temporal dynamics, exhibits inferior performance due to its limited diversity in subgraph samples during training. Furthermore, the method is restricted to predictions for unobserved nodes only. For methods specifically designed for iFun, FUNS's performance is hampered by its ineffective attention mechanism when dealing with unobserved nodes lacking historical data. As an extension of IGNNK, UIGNN incorporates both Kriging and prediction errors to guide model training for forecasting future states of both observed and unobserved nodes. While this approach demonstrates superior performance compared to IGNNK and achieves the second-best results, it shares IGNNK's limitation of inadequate temporal dynamics modeling, thus performing worse than our proposed MoGERNN.

Focusing on the critical task of predicting unobserved nodes within the METR-LA dataset, our method achieves significant reductions in key error metrics relative to the most competitive baseline, with decreases of 16\% in MAPE, 14\% in MAE, and 13\% in RMSE for all prediction horizon in average. Moreover, the prediction accuracy gap between AAS and VS reveals that predicting unobserved locations poses significant challenges. Based on the predictions generated by our model, the error for unobserved locations is 1.2-1.9 times that of observed locations. This substantial difference underscores the complexity inherent in extrapolating to unseen spatial points and highlights the need for advanced methodologies in addressing such prediction tasks.

\cref{fig:map} illustrates the prediction performance of various models during an evening rush hour (May 24, 2012, 17:30) for the METR-LA dataset. The first column depicts the sensor measurements at this time point, while the remaining three columns represent the speed predictions of each model based on historical data. For clarity, prediction results for different sensors are presented in different rows. It is evident that FUNS consistently underperforms compared to the other two models, especially when the speed is lower than 50 mph. UIGNN and MoGERNN showed comparable strong prediction performance at locations where physical sensors are deployed. Our approach, however, shows improved accuracy in both high-speed and low-speed scenarios at unobserved locations. This superiority in performance is notably visible in the results highlighted within the red frames in the third-row subplot.

\begin{figure}[t]
    \centering
    \begin{subfigure}[b]{\textwidth}
        \centering
        \includegraphics[width=\textwidth]{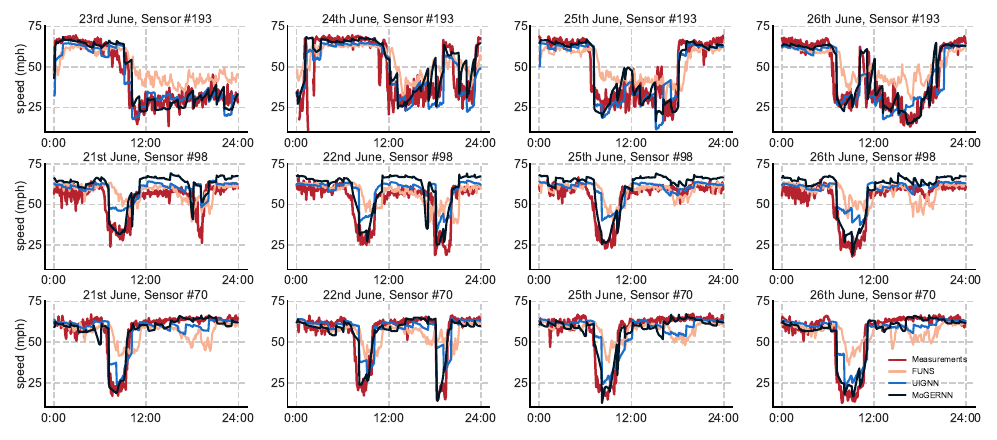}
        \vspace{-2em}
        \caption{Unobserved locations.}
        \label{fig:subfig-a}
    \end{subfigure}
    \quad
    \begin{subfigure}[b]{\textwidth}
        \centering
        \includegraphics[width=\textwidth]{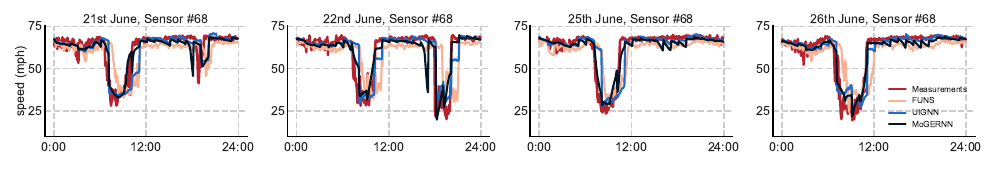}
        \vspace{-2em}
        \caption{Observed location.}
        \label{fig:subfig-b}
    \end{subfigure}
    \vspace{-2em}
    \caption{Predicting results for METR-LA. (a) show the results of unobserved locations, including three virtual sensors (Sensor \#193, \#98, \#70). (b) show the results of observed locations, including one physical sensor (Sensor \#68).}
    \label{fig:SPG}
\end{figure}

Given the paramount importance of congestion evolution in traffic management tasks, the data from METR-LA that exhibit significant congestion characteristics are selected for further analysis. As shown in \cref{fig:SPG}, it illustrates a comparison between observed and predicted time series for selected sensors. Specifically, \cref{fig:subfig-a} displays results for unobserved locations, while \cref{fig:subfig-b} shows results for observed locations. We can find that our MoGERNN demonstrates superior performance in predicting congestion evolution, with this advantage being particularly pronounced at unobserved locations compared to other baselines.

\begin{figure}[t]
    \centering
    \includegraphics[width=1\linewidth]{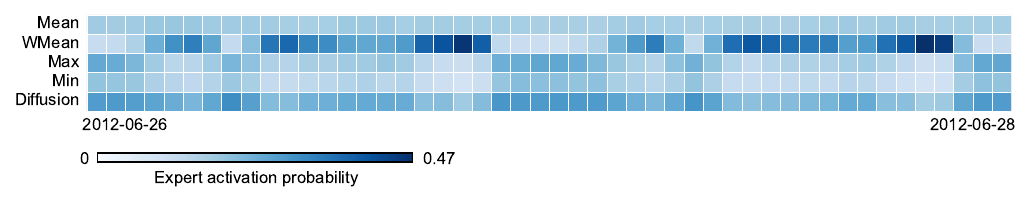}
    \vspace{-2em}
    \caption{Average expert activation probability across all nodes over time for test dataset in METR-LA.}
    \label{fig:test_gate_weights}
\end{figure}
\begin{figure}[t]
    \centering
    \includegraphics[width=1\linewidth]{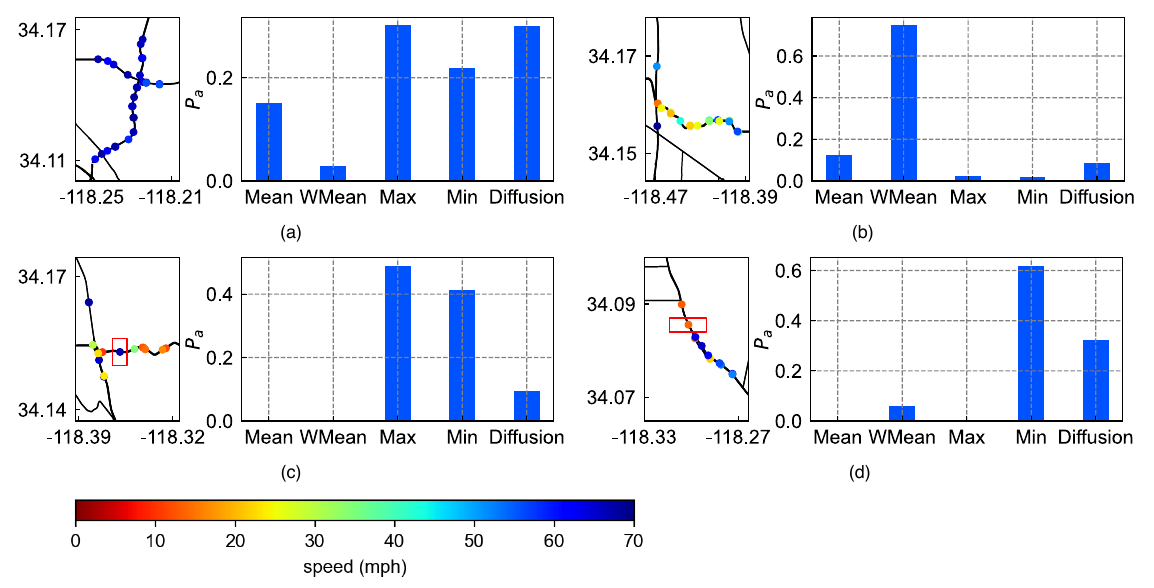}
    \caption{Analysis of expert activation patterns on METR-LA dataset. (a) and (b) Expert activation probabilities in a local region. (c) and (d) Expert activation probability at a single location (red box).}
    \label{fig:detail_gate_weights}
\end{figure}

As previously described, MoGERNN constructs diverse spatial aggregation patterns through multiple graph experts and employs a sparse gating network to dynamically assign and fuse the most suitable experts, thus enhancing the exiting works. To provide insights into this dynamic mechanism, we visualized the expert activation probabilities (aggregated across all nodes) at different time points during the test period from June 26 to June 28, 2012, on the METR-LA dataset, as shown in \cref{fig:test_gate_weights}. The results reveal significant variations in expert activation probabilities across different time points, demonstrating MoGERNN's ability to dynamically adjust spatial modeling strategies based on traffic conditions. In the long-term statistics, the weighted average graph expert exhibits the highest activation probability, which aligns with the fundamental understanding of spatial continuity in traffic flow.

To further illustrate the mechanism of MoGE, we analysis expert activation probabilities in four typical scenarios, as shown in \cref{fig:detail_gate_weights}. \cref{fig:detail_gate_weights}a and \cref{fig:detail_gate_weights}b display the expert activation probability distributions in two local regions, while \cref{fig:detail_gate_weights}c and \cref{fig:detail_gate_weights}d focus on a single location marked by a red box. In the region shown in \cref{fig:detail_gate_weights}a, traffic states exhibit high spatial smoothness, consistent with the "stronger correlation at closer distances" characteristic, leading to the highest activation probability for the weighted average expert. In contrast, the region in \cref{fig:detail_gate_weights}b shows uniformly smooth traffic flow with minimal speed variations, resulting in significantly reduced activation of the weighted average expert while maximum value, minimum value experts, and diffusion models play more crucial roles. In \cref{fig:detail_gate_weights}c, where surrounding nodes experience congestion while the boxed location remains smooth, the activation probabilities of mean and weighted mean experts decrease significantly. The model relies mainly on maximum and minimum experts instead, as the maximum expert captures related smooth traffic states from more distant locations, while the minimum expert provides maximum information gain by capturing the most distinct states.  A similar scenario is observed in \cref{fig:detail_gate_weights}d, where the boxed location experiences low-speed conditions, and the model primarily activates the minimum expert to capture the relevant congested nodes in the surrounding area. These observations demonstrate how MoGERNN flexibly adjusts its spatial modeling strategy across different traffic scenarios for accurate prediction.

\subsection{Adaptability to the structural changes of sensing networks (Q2)} \label{sec:q2}

\begin{table}[t]
  \centering
   \caption{Sensor configuration to simulate the structural changes of sensing network.}
  \label{tab:sc}
  \vspace{-0.7em}
   \resizebox{\columnwidth}{!}{%
   \begin{tabular}{ccccc}
    \toprule
    \textbf{Dataset} & \textbf{Always Available Sensors (AAS)}  & \textbf{Virtual Sensors (VS)} & \textbf{New Added Sensors (NAS)} & \textbf{Failed Sensors (FS)} \\
    \hline
    METR-LA & 137  & 30  & 20 &  20\\
    PEMS-BAY & 196  & 65 & 32 & 32\\
    \bottomrule
   \end{tabular}}
\end{table}

To simulate the structural changes in sensor network, we extend the sensor configuration from \cref{sec:q1} by introducing NS and FS. \cref{tab:sc} presents the quantities of all four sensor types under this new configuration. Given the aims to evaluate the model's adaptability to changes in sensor configurations, we directly apply the model trained on the sensor configuration from Q1 to the current sensing environment for testing. In addition, to comprehensively assess the model's adaptability to the changes of sensor networks, we also compare its performance with that of a model retrained on the new sensor network configuration. The average prediction performance of all prediction horizons is presented in \cref{fig:PPUD}. For brevity, we present only the results of UIGNN and FUNS for comparison. The results demonstrate that our model outperforms these baselines in the majority of scenarios across all four settings. This advantage is particularly pronounced in VS and FS scenarios, highlighting MoGERNN's superior spatial extrapolation capabilities. Moreover, despite the absence of NAS training data, MoGERNN achieves comparable performance between NAS and AAS, indicating satisfactory zero-shot prediction capability.

In comparing the performance of MoGERNN with its retrained counterpart, shown in \cref{fig:retrained}, we observe that the retrained version achieves slightly better predictive accuracy. However, the moderate performance gap suggests that MoGERNN maintains robust predictive capability even without retraining. This adaptability to different sensor configurations offers practitioners the flexibility to make retraining decisions based on operational requirements and resource constraints, thereby enhancing the model's practical utility in real-world applications.

\begin{table}[t]
\caption{Average prediction performance of all prediction horizons under the structural changes in sensing network.}
\label{fig:PPUD}
\vspace{-0.7em}
\centering
\resizebox{0.8\textwidth}{!}{%
\begin{tabular}{llcccccc}
\toprule
\multirow{2}{*}{\textbf{Sensor}} & \multirow{2}{*}{\textbf{Models}} & \multicolumn{3}{c}{\textbf{METR-LA}}           & \multicolumn{3}{c}{\textbf{PEMS-Bay}}         \\ \cline{3-8} 
                                 &                                  & \textbf{MAPE}  & \textbf{MAE}  & \textbf{RMSE} & \textbf{MAPE} & \textbf{MAE}  & \textbf{RMSE} \\ \hline
\multirow{4}{*}{Virtual sensor}  & UIGNN                            & {\ul16.13}     & {\ul6.45}     & {\ul9.74}     & {\ul8.71}     & {\ul3.90}     & {\ul6.28}    \\
                                 & FUNS                             & 20.53          & 7.60          & 10.88         & 10.60         & 4.67          & 7.61          \\
                                 & MoGERNN (ours)                   & \textbf{14.77} & \textbf{6.01} & \textbf{8.92} & \textbf{7.73} & \textbf{3.59} & \textbf{5.94} \\ \cline{2-8} 
                                 & Improve (\%)                     & +8.43          & +6.82         & +8.42         & +11.25         & +7.95         & +5.41         \\ \hline
\multirow{4}{*}{Always available sensor}          & UIGNN           & {\ul10.80}     & {\ul3.92}     & {\ul6.96}     & {\ul 4.98}    & {\ul2.17}     & {\ul4.44}    \\
                                 & FUNS                             & 16.17          & 5.60          & 9.08          & 7.71          & 3.18          & 6.39          \\
                                 & MoGERNN (ours)                   & \textbf{10.56} & \textbf{3.64} & \textbf{6.59} & \textbf{4.60} & \textbf{1.94} & \textbf{4.15} \\ \cline{2-8} 
                                 & Improve (\%)                     & +2.22          & +7.14         & +5.32         & +7.63         & +10.60         & +6.53          \\ \hline
\multirow{4}{*}{New added sensor}                 & UIGNN           & \textbf{11.16}     & \textbf{4.00} & {\ul7.12}     & {\ul 5.11}    & {\ul 2.12}    & {\ul 4.50}    \\
                                 & FUNS                             & 16.13          & 5.66          & 9.91          & 7.74          & 3.05          & 6.31          \\
                                 & MoGERNN (ours)                   & {\ul11.38} & {\ul 4.04}    & \textbf{6.94} & \textbf{5.02} & \textbf{2.08} & \textbf{4.54} \\ \cline{2-8} 
                                 & Improve (\%)                     & -1.97          & -1.00         & +2.53         & +1.76         & 1.87          & -0.09          \\ \hline
\multirow{4}{*}{Failed sensor}   & UIGNN                            & {\ul18.50}     & {\ul6.40}     & {\ul9.82}     & {\ul9.27}     & {\ul3.80}     & {\ul6.40}    \\
                                 & FUNS                             & 22.31          & 8.02          & 11.28         & 11.38         & 4.52          & 7.70          \\
                                 & MoGERNN (ours)                   & \textbf{13.24} & \textbf{5.00} & \textbf{7.78} & \textbf{7.27} & \textbf{3.12} & \textbf{5.49} \\ \cline{2-8} 
                                 & Improve (\%)                     & +28.43         & +21.88        & +20.77        & +21.57        & +17.89        & +14.22        \\
\bottomrule
\end{tabular}}
\end{table}

\begin{table}[t]
\caption{Performance comparison of MoGERNN with its retrained counterpart after sensor change.}
\label{fig:retrained}
\vspace{-0.7em}
\centering
\resizebox{0.8\textwidth}{!}{%
\begin{tabular}{llcccccc}
\toprule
\multirow{2}{*}{\textbf{Sensor}} & \multirow{2}{*}{\textbf{Models}} & \multicolumn{3}{c}{\textbf{METR-LA}}         & \multicolumn{3}{c}{\textbf{PEMS-Bay}}        \\ \cline{3-8} 
                                 &                                  & \textbf{MAPE} & \textbf{MAE} & \textbf{RMSE} & \textbf{MAPE} & \textbf{MAE} & \textbf{RMSE} \\ \hline
\multirow{2}{*}{Virtual sensor}  & MoGERNN                          & 14.77         & 6.01         & 8.92          & 7.73          & 3.59         & 5.94          \\
                                 & MoGERNN-retrained                & 14.51         & 5.92         & 9.74          & 7.93          & 3.60         & 6.05          \\ \hline
\multirow{2}{*}{Always available sensor}          & MoGERNN         & 10.56         & 3.64         & 6.59          & 4.60          & 1.94         & 4.15          \\
                                 & MoGERNN-retrained                & 10.02         & 3.43         & 6.26          & 4.44          & 1.90         & 4.00          \\ \hline
\multirow{2}{*}{New added sensor}                 & MoGERNN         & 11.01         & 4.04         & 6.94          & 5.02          & 2.08         & 4.54          \\
                                 & MoGERNN-retrained                & 10.81         & 3.83         & 6.30          & 4.34          & 1.84         & 4.02          \\ \hline
\multirow{2}{*}{Failed sensor}                    & MoGERNN         & 13.24         & 5.00         & 7.78          & 7.27          & 3.12         & 5.49          \\
                                 & MoGERNN-retrained                & 12.69         & 4.45         & 7.24          & 6.79          & 2.93         & 5.14          \\ 
\bottomrule
\end{tabular}}
\end{table}

\subsection{Impact of the VS-to-AAS Ratio (Q3)}\label{sec:q3}

\begin{figure}[t]
    \centering
    \includegraphics[width=1\linewidth]{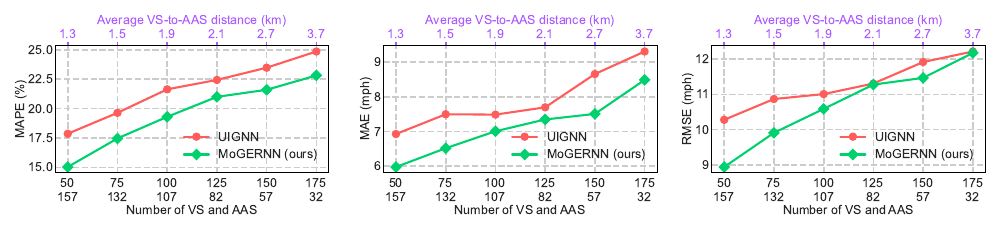}
    \vspace{-2em}
    \caption{Model performance under different ratios of VS to AAS. The first row of x-ticks labels indicates the number of VS, while the second row shows the number of AAS.}
    \label{fig:mpuvs}
\end{figure}

To examine the impact of the VS-to-AAS ratio on model performance, we randomly sample 50 to 175 VS from the METR-LA dataset. For simplicity, NAS and FS are not considered in this analysis. The performance trends of the model as the VS-to-AAS ratio changes are illustrated in \cref{fig:mpuvs}. In addition to the VS-to-AAS ratio, we introduce an additional metric, Average VS-to-AAS Distance $D_{v2a}$, to better characterize the spatial extrapolation capability of the model, as shown in purple ticks in the \cref{fig:mpuvs}. The definition of  $D_{v2a}$ is as follows: 
\begin{equation}
    d_{min}^i = \min_{j:j \in \text{AAS, } i \neq j}d_{i,j}
\end{equation}
\begin{equation}
    D_{v2a} = \operatorname{Avg}_{i \in \text{VS}} d_{min}^i
\end{equation}

According to \cref{fig:mpuvs}, it is evident that our method consistently outperforms our main competitor, UIGNN, across various VS-to-AAS ratios. Given the same level of predictive performance, our approach typically requires at least 25 fewer sensors than UIGNN. Furthermore, based on the MAE and \(D_{v2a}\) values, we can observe that the spatial extrapolation capability of our model is approximately 1.4 times that of UIGNN.

\begin{table}[t]
\caption{Effect of Encoder-Decoder and MoGE module.}
\vspace{-0.7em}
\centering
\label{tab:ablation}
\resizebox{0.8\columnwidth}{!}{%
\begin{tabular}{lcccccc}
\toprule
\multirow{2}{*}{Models}                   & \multicolumn{3}{c}{Always available sensor (AAS)}  & \multicolumn{3}{c}{Virtual sensor (VS)}   \\ \cline{2-7} 
                                          & MAPE   & MAE    & RMSE   & MAPE   & MAE    & RMSE   \\ \hline
w/o ED                                    & 11.63  & 4.15   & 7.32   & 18.13  & 6.66   & 9.82   \\
MoGERNN (ours)                            & 10.24  & 3.57   & 6.44   & 15.03  & 5.98   & 8.94   \\ \cline{2-7} 
Improve (\%)                              & +11.95 & +13.98 & +12.02 & +17.10 & +10.21 & +8.96  \\ \hline
w/o MoGE                                  & 9.93   & 3.58   & 6.42   & 16.14  & 6.77   & 9.98   \\
MoGERNN (ours)                            & 10.24  & 3.57   & 6.44   & 15.03  & 5.98   & 8.94   \\ \cline{2-7} 
Improve (\%)                              & -3.12  & +0.28  & -0.31  & +6.88  & +11.67 & +10.42 \\ \hline
replace MoGE   with mean                  & 10.47  & 3.61   & 6.55   & 16.31  & 6.50   & 9.62   \\
MoGERNN (ours)                            & 10.24  & 3.57   & 6.44   & 15.03  & 5.98   & 8.94   \\ \cline{2-7} 
Improve (\%)                              & +2.20  & +1.11  & +1.68  & +7.85  & +8.00  & +7.07  \\ \hline
replace MoGE   with weighted-mean         & 10.31  & 3.60   & 6.48   & 15.98  & 6.28   & 9.33   \\
MoGERNN (ours)                            & 10.24  & 3.57   & 6.44   & 15.03  & 5.98   & 8.94   \\ \cline{2-7} 
Improve (\%)                              & +0.68  & +0.83  & +0.62  & +5.94  & +4.78  & +4.18  \\ \hline
replace MoGE   with max-pooling           & 10.79  & 3.63   & 6.61   & 15.75  & 6.11   & 9.32   \\
MoGERNN (ours)                            & 10.24  & 3.57   & 6.44   & 15.03  & 5.98   & 8.94   \\ \cline{2-7} 
Improve (\%)                              & +5.10  & +1.65  & +2.57  & +4.57  & +2.13  & +4.08  \\ \hline
replace MoGE with min-pooling             & 10.89  & 3.67   & 6.62   & 16.11  & 6.17   & 9.42   \\
MoGERNN (ours)                            & 10.24  & 3.57   & 6.44   & 15.03  & 5.98   & 8.94   \\ \cline{2-7} 
Improve (\%)                              & +5.97  & +2.72  & +2.72  & +6.70  & +3.08  & +5.10  \\ \hline
replace MoGE with DGCN                    & 10.43  & 3.63   & 6.49   & 16.04  & 6.41   & 9.43   \\
MoGERNN (ours)                            & 10.24  & 3.57   & 6.44   & 15.03  & 5.98   & 8.94   \\ \cline{2-7} 
Improve (\%)                              & +1.82  & +1.65  & +0.77  & +6.30  & +6.71  & +5.20  \\ \hline
replace sparse   gating with average sums & 9.96   & 3.58   & 6.45   & 15.97  & 6.81   & 9.82   \\
MoGERNN (ours)                            & 10.24  & 3.57   & 6.44   & 15.03  & 5.98   & 8.94   \\ \cline{2-7} 
Improve (\%)                              & -2.81  & +0.28  & +0.16  & +5.89  & +12.19 & +8.96  \\ 
\bottomrule
\end{tabular}}
\end{table}

\subsection{Ablation Study (Q4)} \label{sec:q4}

To evaluate the effectiveness of the Encoder-Decoder and MoGE in our model, we conduct the following ablation studies, with results consolidated in \cref{tab:ablation}. For simplicity, we present only the average results across all prediction horizons on the METR-LA dataset, without consideration of NAS and FS. The sensor type configuration remains consistent with that in \cref{sec:q1}.

1) Removal of the Encoder-Decoder Module: The first experiment involves removing the encoder-decoder module. Results indicate that this module positively influences predictions for both AAS and VS.

2) Removal of the MoGE Module: In the second experiment, we removed the MoGE module. while AAS performance shows modest improvement with a 3.2\% reduction in MAPE, the VS prediction accuracy substantially deteriorates, exhibiting increases of 6.9\% in MAPE, 10.2\% in MAE, and 8.96\% in RMSE. These results demonstrate that the MoGE module plays a crucial role in maintaining the model's capability to accurately predict at unobserved locations, despite a slight trade-off in performance at observed nodes. This trade-off stems from the removal of self-loops in MoGE's graph expert operations (reflected in \cref{eq:no-self}), which forces nodes to construct their representations primarily through neighboring nodes' features rather than their own historical data. While this design choice is beneficial for unobserved nodes that lack historical information, it slightly compromises the performance at observed nodes where self-historical information would naturally be most relevant for prediction. A potential improvement could be adopting a multitask learning framework that treats observed and unobserved node prediction as separate but related tasks, allowing the model to better balance the performance between these two scenarios.

3) Replace MoGE with Single Graph Expert: To further investigate the role of multiple graph experts in the MoGE module, we conducted experiments retaining only one expert at a time. In this ablation setting, the model degenerates to a static graph structure without the ability to dynamically combine different spatial aggregation patterns. As shown in Table 6, the weighted-mean expert delivers the best performance for AAS predictions, whereas the max-pooling expert excels in VS predictions. MoGERNN, leveraging the strengths of multiple graph experts, achieves superior performance in both AAS and VS than a single graph expert. This confirms the importance of multi-graph message aggregators in the iFun problem.

4) Replace Sparse Gating With Average Sums: To validate the efficacy of using sparse gating networks, we replaced them with simple averaging. While average sum provides some improvement in AAS, it resulted in a noticeable decline in VS predictions compared to the max-pooling expert. In contrast, MoGERNN with sparse gating improved performance in both metrics. Although the MAPE for AAS with sparse gating was slightly higher than the average sum approach (-2.8\% in MAPE), the VS prediction error decreased by 5.9\%, 12.2\% and 9.0\% in MAPE, MAE and RMSE, respectively. This experiment confirms that there is value in designing advanced methods for integrating multi-graph experts.

\begin{figure}[t]
    \centering
    \includegraphics[width=0.9\linewidth]{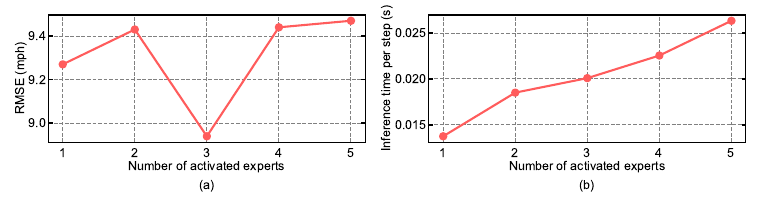}
    \caption{Impact of number of activated experts:(a) performance of state prediction on unobserved nodes in METR-LA dataset, (b) inference time the model.}
    \label{fig:expert_sense}
\end{figure}

\subsection{Parameters study (Q5)} \label{sec:q5}

In this experiment, we investigate the impact of varying numbers of active experts on both model performance and inference time. We conduct experiments on the METR-LA dataset, maintaining the same sensor type configuration as in \cref{sec:q1}. As shown in \cref{fig:expert_sense}, subplot (a) presents the RMSE on unobserved nodes across all prediction horizon, while subplot (b) illustrates the inference time. The results demonstrate a U-shaped performance pattern as the number of active experts increases, with optimal performance achieved when activating three experts. Notably, the performance with just one activated expert is comparable to that with three experts. Referring to the results in \cref{tab:PP}, this single-expert performance also surpasses the baseline methods, demonstrating the effectiveness of our expert design. Meanwhile, the runtime increases linearly with the number of activated experts, aligning with expectations. Also, the inference time increase caused by multiple experts is acceptable- from activating a single expert to all five experts, the total inference time only increases from 0.014 s to 0.026 s, which is well within the requirements for real-time practice.

\section{Conclusion} \label{sec:con}

In this paper, we propose MoGERNN, an inductive spatio-temporal graph learning framework that enables traffic state forecasting at unobserved locations. Our experiments demonstrate that the proposed method consistently outperforms four baseline models for both observed and unobserved nodes. For congestion states, our model accurately predicts congestion evolution based on historical data, even in areas without sensors. This provides richer and more effective information for traffic management and control compared to traditional traffic predictors. Additionally, our method adapts more effectively to the structural changes of sensor network, such as the addition or failure of sensors compared to baselines. Even when compared to MoGERNN retrained with new sensor configurations, the model trained on historical sensor configuration maintains competitive performance. Notably, the traffic state prediction accuracy for locations with newly added sensors closely matches that of locations with continuous data, highlighting our method's superior zero-shot prediction capability.

Several avenues for future research merit exploration. First, uncertainty estimation: the quantification of uncertainty is of paramount importance \citep{Wen2023Diff}, particularly given that these predictions inform critical traffic management strategies. The need for this quantification becomes even more pronounced when dealing with unobserved nodes. Given that the spatial extrapolation capabilities of models are limited, arbitrary application to unobserved nodes may lead to erroneous decisions  in traffic management. Different from nodes with observed data, model errors at unobserved nodes cannot be evaluated, so uncertainty quantification can assist us in understanding model performance at these locations. Second, sensor deployment analysis: identifying the relationship between sensor deployment patterns and model performance would help establish clear guidelines for practical applications. This includes understanding what types of sensor distributions allow reliable predictions and what patterns lead to degraded performance. Third, multi-source data fusion: the iFun problem can be ill-posed when the existing sensor deployment is too sparse, in which case we would like the model to be able to incorporate other available data sources to solve the iFun, e.g., connected vehicle data \citep{Xu2024multisource}. With the support of multi-source data, we can also attempt to address more valuable and challenging problems, such as traffic flow estimation and prediction at unobserved locations \citep{nie2023flowestimation}. Third, multivariate data processing capability: typically, fixed traffic sensors simultaneously measure three key variables: speed, flow, and occupancy. By extending the model to accommodate multivariate data, we could not only fully utilize the available data but also potentially enhance the model by introducing regularization constraints based on the prior dependencies among the multivariate variables, such as the fundamental diagram.

\section*{CRediT authorship contribution statement}

\textbf{Qishen Zhou}: Conceptualization, Methodology, Software, Investigation, Formal analysis,
Writing – original draft. \textbf{Yifan Zhang}: Conceptualization, Formal analysis, Writing – review \&
editing. \textbf{Michail A. Makridis}: Supervision, Writing – review \& editing. \textbf{Anastasios Kouvelas}:
Supervision, Writing – review \& editing. \textbf{Yibing Wang}: Supervision, Writing – review \& editing.
\textbf{Simon Hu}: Funding acquisition, Resources, Supervision, Writing – review \& editing.

\section*{Acknowledgements}
The authors would like to thank the anonymous reviewers for their valuable comments and suggestions that have significantly improved this paper. This work was supported in part by ``Pioneer'' and ``Leading Goose'' R \& D Program of Zhejiang (2023C03155), the National Key R\&D Program of China (2023YFB4302600), the National Natural Science Foundation of China (52131202, 72350710798), the Smart Urban Future (SURF) Laboratory, Zhejiang Province, Zhejiang University Global Partnership Fund, and Zhejiang University Sustainable Smart Livable Cities Alliance (SSLCA) led by Principal Supervisor Simon Hu, and the Provincial Key R\&D Program of Zhejiang (2024C01180; 2022C01129), the National Natural Science Foundation of China (52272315), Ningbo International Science and Technology Cooperation Program (2023H020) led by co-supervisor Yibing Wang. Qishen Zhou acknowledges support from the China Scholarship Council for his one-year visiting at ETH Zurich.

\bibliographystyle{elsarticle-harv} 
\bibliography{cas-refs}

\end{document}